\pdfoutput=1

\documentclass[11pt]{article}
\usepackage{xcolor}
\usepackage{amsmath}
\usepackage{amssymb}
\usepackage{xspace}

\usepackage{acl}
\usepackage{times}
\usepackage{latexsym}
\usepackage{scalerel}
\usepackage{subcaption}
\usepackage{float}

\usepackage[T1]{fontenc}
\usepackage[utf8]{inputenc}
\usepackage{xfrac}

\usepackage{cleveref}
\crefname{section}{\S}{\S\S}
\Crefname{section}{\S}{\S\S}
\crefname{table}{Table}{}
\crefname{figure}{Fig.}{Figs.}
\crefname{algorithm}{Algorithm}{}
\crefname{equation}{eq.}{}
\crefname{appendix}{App.}{}
\crefname{thm}{Theorem}{}
\crefname{prop}{Proposition}{}
\crefname{cor}{Corollary}{}
\crefname{observation}{Observation}{}
\crefname{assumption}{Assumption}{}
\crefformat{section}{\S#2#1#3}
\crefformat{footnote}{#2\footnotemark[#1]#3}

\newcommand{\xx}{\mathbf{x}}
\newcommand{\yy}{\mathbf{y}}
\newcommand{\vocab}{\mathcal{V}}
\newcommand{\vocabeos}{\overline{\vocab}}
\newcommand{\eos}{\textsc{eos}\xspace}
\newcommand{\vtheta}{{\boldsymbol \theta}}
\newcommand{\model}{p_{\scaleto{\vtheta}{4pt}}}

\newcommand{\bleu}{\textsc{bleu}\xspace}
\newcommand{\zerovec}{{\footnotesize $\overset{\rightarrow}{0}$}}
\newcommand{\pW}{p_{\scaleto{\bW\bphi}{4pt}}}
\newcommand{\defeq}[0]{\mathrel{\stackrel{\textnormal{\tiny def}}{=}}}
\newcommand{\bW}{\mathbf{W}}
\newcommand{\bb}{\mathbf{b}}
\newcommand{\nbb}{\mathrm{b}}
\newcommand{\bphi}{\boldsymbol{\phi}}
\newcommand{\repfunction}{\bphi(\yy_{<t})}

\usepackage[disable,textsize=tiny]{todonotes}
\newcommand{\note}[4][]{\todo[author=#2,color=#3,size=\scriptsize,fancyline,caption={},#1]{#4}} %
\newcommand{\tiago}[2][]{\note[#1]{tiago}{cyan!40}{#2}}

\newcommand{\clara}[2][]{\note[#1]{clara}{yellow!40}{#2}}

\newcommand{\ucambridge}{3}
\newcommand{\deepmind}{2}
\newcommand{\ethz}{1}
\usepackage{inconsolata}

\newcommand{\suggest}[2]{{\color{cyan}#2}}

\title{A Natural Bias for Language Generation Models}

\author{
 Clara Meister\thanks{\,\,\,Work done during internship at DeepMind.}$^{\,\,\,,\ethz}$,~\;~Wojciech Stokowiec$^{\deepmind}$,~\;~Tiago Pimentel$^{\ucambridge}$, \\
\textbf{Lei Yu$^{\deepmind}$,~\;~ Laura Rimell$^{\deepmind}$,~\;~Adhiguna Kuncoro$^{\deepmind}$} \\
  $^{\ethz}$ETH Z\"{u}rich~\;~ $^{\deepmind}$DeepMind~\;~ $^{\ucambridge}$University of Cambridge \\
  \texttt{\href{mailto:meistecl@inf.ethz.ch}{meistecl@inf.ethz.ch}}~\;~ \texttt{\href{mailto:wstokowiec@deepmind.com}{wstokowiec@deepmind.com}}~\;~ \texttt{\href{mailto:tp472@cam.ac.uk}{tp472@cam.ac.uk}} \\
  \texttt{\href{mailto:leiyu@deepmind.com}{leiyu@deepmind.com}}~\;~ \texttt{\href{mailto:laurarimell@deepmind.com}{laurarimell@deepmind.com}}~\;~ \texttt{\href{mailto:akuncoro@deepmind.com}{akuncoro@deepmind.com}}
  }

\begin{document}
\maketitle
\begin{abstract}

After just a few hundred training updates, a standard probabilistic model for language generation has likely not yet learnt many semantic or syntactic rules of natural language, making it difficult to estimate the probability distribution over next tokens. Yet around this point, these models have identified a simple, loss-minimising behaviour: to output the unigram distribution of the target training corpus. The use of such a heuristic raises the question: Can we initialise our models with this behaviour and save precious compute resources and model capacity? Here we show that we can effectively endow standard neural language generation models with a separate \emph{module} that reflects unigram frequency statistics as \emph{prior knowledge}, simply by initialising the bias term in a model's final linear layer with the log-unigram distribution. We use  neural machine translation as a test bed for this simple technique and observe that it: (i) improves learning efficiency; (ii) achieves better overall performance; and perhaps most importantly (iii) appears to disentangle strong frequency effects by encouraging the model to specialise in non-frequency-related aspects of language.\looseness=-1

\end{abstract}

\section{Introduction}

Consider the structure of a number of core tasks in natural language processing (NLP): predicting the next word following a given context. 
What if you did not understand the context -- for example, if you did not know the language? In the absence of such knowledge, the optimal prediction would be the language's most frequent word. 
In fact, optimally one would predict each word according to its (unigram) frequency.\footnote{Notably, void of contextual clues, models of human language processing \cite{morton1969interaction} would default to similar strategies. A word's frequency also influences its age of acquisition \cite{gilhooly1980age,catriona1997age}, and the time taken to produce it in speech \cite{effects_1998,ZEVIN20021}.}
This is precisely the strategy that neural language models have been empirically observed to employ during early training stages \cite{chang-bergen-2022-word} -- before they have learnt a language's syntax or semantics.\looseness=-1 

\begin{figure}
    \centering
    \includegraphics[width=\columnwidth]{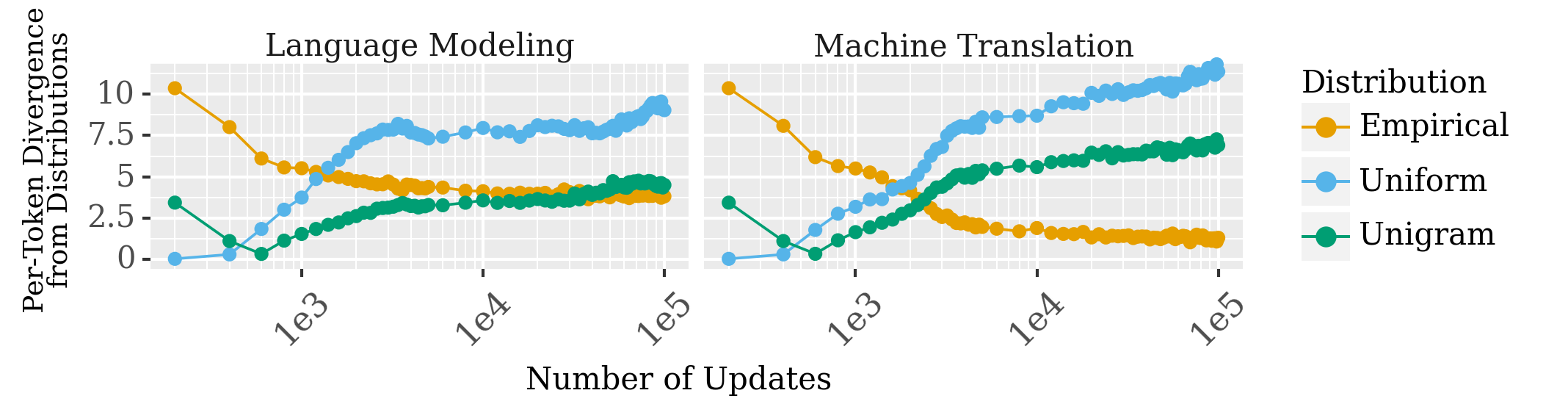}
    \caption{Average per-token divergence of the model from unigram, uniform, and empirical distributions of respective training set as a function of training step (log-scale). Early in training, the model output closely matches the unigram distribution for all contexts. \looseness=-1 }
    \label{fig:unigram_div}
\end{figure} 

Although this strategy of predicting the unigram distribution emerges early in training, it still takes the model hundreds (or even thousands) of parameter updates to learn it from a cold start %
\citep[see \cref{fig:unigram_div} or][ Fig. 5]{chang-bergen-2022-word}.
Yet a straightforward factorisation of a language model's final linear layer shows that we can in fact encode this frequency-related knowledge \emph{prior to any optimisation},\footnote{The unigram distribution of the training data is known before optimisation, as it is often computed when building vocabularies or tokenising; hence this approach should come at no extra cost.} with the goal of bypassing this early stage of learning:
Concretely, this is done by setting the bias term in a model's final linear layer to the log-unigram distribution of the training data. Mathematically, this setup can be loosely interpreted as a modular ``product of experts'' \citep{hinton}, where the bias term represents a simple unconditional distribution over the vocabulary, thus allowing the input-dependent logits to specialise in capturing contextual information.
Indeed, we argue that a more modular design that disentangles word-frequency effects from contextual information may be desirable, given the recently-observed negative effects of %
word frequency statistics on models' generalisation abilities 
\cite{wei-etal-2021-frequency,outlier_dims,rajaee-pilehvar-2022-isotropy}.\looseness=-1

While this initialisation approach has been historically used in language models \cite[\emph{inter alia}]{mnih_graphical,pmlr-v32-botha14,7277009}, it has not seen widespread adoption within our current language generation architectures -- an observation we attribute to uncertainty around whether the bias term automatically specialises to capture frequency  without explicit encouragement to do so. We first observe that this is \emph{not} the case -- in fact, the final-layer bias term rarely changes from its \emph{random} initialisation (see \cref{app:bias_change}), suggesting frequency is encoded elsewhere in the model parameters. 
We then empirically explore the impact of this initialisation on various aspects of model behaviour \suggest{}{--} within the context of current Transformer models for machine translation \suggest{}{--} including overall performance, learning efficiency, and the relationship between model-assigned probability and word frequency. %
We find this initialisation indeed leads to increased training efficiency: models achieve higher \bleu scores earlier on in training. More surprisingly, it also leads to improved \emph{overall performance}. We discuss several potential reasons for these results, including changes to training dynamics and a mitigation of overfitting to surface statistics.\looseness=-1 %
\section{Probabilistic Language Generators}

\subsection{Preliminaries}
We consider neural probabilistic models $\model$ for language generation. While there are a variety of architectural choices that can be made, most are autoregressive and follow a local-normalisation scheme. Explicitly,  given prior context $\yy_{<t}\defeq \langle y_0, \dots, y_{t-1}\rangle$, these models output a probability distribution $\model(\cdot \!\mid\! \yy_{<t})$ over the next token $y \in \vocabeos \defeq \vocab \cup \{\eos\}$, where $\vocab$ is the model's predefined vocabulary and \eos is a special end-of-sequence token. %
To ensure that $\model$ provides a valid probability distribution, the output of the model is projected onto the probability simplex $\Delta^{|\vocabeos|-1}$ using a softmax transformation after a (learnt) linear projection layer:\footnote{While $\bphi$ is also conditioned on a source sentence $\xx$ in the case of machine translation, we leave this implicit in our equations for notational simplicity.}\looseness=-1
\begin{align}
    \model(y \!\mid\! \yy_{<t}) &= \mathrm{softmax}\left(\bW\, \repfunction + \bb\right)_y \\
    &\defeq
    \frac{e^{\bW_y\, \repfunction + \nbb_{y}}}{\sum_{y'\in \vocabeos}e^{\bW_{y'}\, \repfunction + \nbb_{y'}}} \label{softmax}
\end{align}
where  $\bW \in \mathbb{R}^{|\vocabeos|\times d}$ denotes a weight matrix,  $\bb \in  \mathbb{R}^{|\vocabeos|}$ a bias vector, and  $\bphi: \vocab^* \rightarrow \mathbb{R}^{d}$ the model's $d$-dimensional encoding for a given context.\footnote{We index vectors and matrices using $y$, assuming an isomorphic mapping between $y\in\vocabeos$ and integers $[1, \dots, |\vocabeos|]$.}

A number of prior studies have investigated whether -- and if so, at what stage during the learning process -- NLP models learn various linguistic phenomena \citep[][\emph{inter alia}]{alain2016understanding,adi2016fine}. 
Among the key findings are that language models reflect the statistical tendencies exhibited by their respective training corpora \cite{takahashi_statistical,takahashi_evaluating,meister+al.acl21}; some of which are learnt early on in training \citep{liu-etal-2021-probing-across}. 
For example, \citet{chang-bergen-2022-word} observe 
that, after only $\sim 1000$ training updates, language models' outputs are approximately equal to the unigram distribution, regardless of the context that they condition on.
We similarly observe this for machine translation models (see \cref{fig:unigram_div}).

\subsection{A Natural Bias}
These learning trends motivate trying to supply language generation models with a natural starting point: the unigram distribution. 
Fortunately, this form of prior knowledge can be modularly encoded in standard neural models using the bias term of the final, pre-softmax linear layer. Consider the standard operation for projecting the output of the model onto the probability simplex. Upon closer inspection, we see that \cref{softmax} has an interpretation as the product of two probability distributions, up to a normalisation constant:
\begin{align}
    \model(\cdot \!\mid\! \yy_{<t}) &\propto e^{\bW\, \repfunction}  \cdot e^{\bb} \\
    &\propto \pW(\cdot \!\mid\! \yy_{<t}) \cdot p_b(\cdot)\label{eq:breakdown}
\end{align}
i.e., one described by $\pW(\cdot \!\mid\! \yy_{<t})$ -- which is \emph{contextual} as it depends on the input $\yy_{<t}$ -- and a separate, \emph{non-contextual} term denoted by $p_b(\cdot)$. Thus, we can qualitatively view this setup as factorising the model's prediction into these two components.\footnote{Given this decomposition, one might expect that models learn to use the bias term to encode frequency on their own. Yet we do not find this to be the case empirically (\cref{app:bias_change}).\looseness=-1} 
In this light, it makes intuitive sense that $p_b$ should be the unigram distribution -- a distribution which optimally predicts (w.r.t.  negative log-likelihood loss) the next-token when there is no contextual information to condition on.
Note that such a setup -- where a probability distribution is modelled using a product of several simpler distributions, each of which can specialise on modelling one aspect of the problem -- is referred to as a product of experts \cite{hinton}.\footnote{The comparison of mixtures and products of experts is well summarised by the phrase: a single expert in a mixture has the power to pass a bill while a single expert in a product has the power to veto it. Each paradigm has its advantages. Here, we argue that the latter is more suitable for language modelling, as the mixture formulation presents the issue that high-frequency tokens will be strongly ``up-voted'' by the expert corresponding to the unigram distribution. As these models already have a propensity to select high frequency tokens, even in improper contexts \cite{wei-etal-2021-frequency}, this is arguably an undesirable trait.\looseness=-1 }

\section{Related Work}\label{sec:related_work}
As previously mentioned, prior work has likewise taken advantage of the interpretation of the bias term  as a frequency offset when initialising model parameters \cite[\emph{inter alia}]{mnih_graphical,pmlr-v32-botha14,7277009}. 
Yet such techniques  have fallen to the wayside for a number of years now, as other more prominent determinants of model performance and training efficiency have dominated the community's attention. We revisit this initialisation strategy in the context of  today's neural language  models. 

The practice of directly incorporating unigram probabilities into next-word predictions can be likened to the back-off methods proposed in the $n$-gram literature \cite{479394,CHEN1999359}.\footnote{How to properly estimate the unigram distribution itself is an important, but often overlooked, question. In our work, we consider a predefined and finite vocabulary $\vocab$; and estimate probabilities using their frequency in a training corpus. For a longer discussion on this see \citet{nikkarinen+al.acl-findings21}.} Indeed, there is an entire class of methods built around learning deviations from some base reference distribution, some of which have been employed specifically for language modelling \cite{berger-printz-1998-comparison,teh-2006-hierarchical,grave2017improving}. More recently, \citet{ngram_residual} cast neural language modelling as the learning of the residuals not captured by $n$-gram models and \citet{baziotis-etal-2020-language} use language models as a reference distribution for training low resource machine translation models. %

Another class of prior work has similarly explored efficient strategies for model weight initialisation \citep[][\emph{inter alia}]{pmlr-v9-glorot10a,Le2015ASW,reinitialization}, including random variable choices and re-initialisation criterion. In a similar vein, \citet{ben-zaken-etal-2022-bitfit} investigate the usefulness of the bias term, albeit for efficient fine-tuning techniques. They show that, often, modifying solely the bias parameters during fine-tuning provides comparable performance to updating the entire model. Both our results thus showcase the usefulness of this simple set of parameters for natural language processing tasks. %

Other works have also embraced frameworks akin to product or mixture of experts  in language modelling or generation tasks. For example \citet{neubig-dyer-2016-generalizing} combine neural and count-based language models in a mixture of experts paradigm; \citet{artetxe-etal-2022-efficient} take advantage of the mixture of experts structure to propose a compute-efficient language modelling architecture.    
In contrast, we suggest a simple initialisation method that does not require training additional models or major changes to model architectures.

\begin{figure}
    \centering
    \includegraphics[width=\columnwidth]{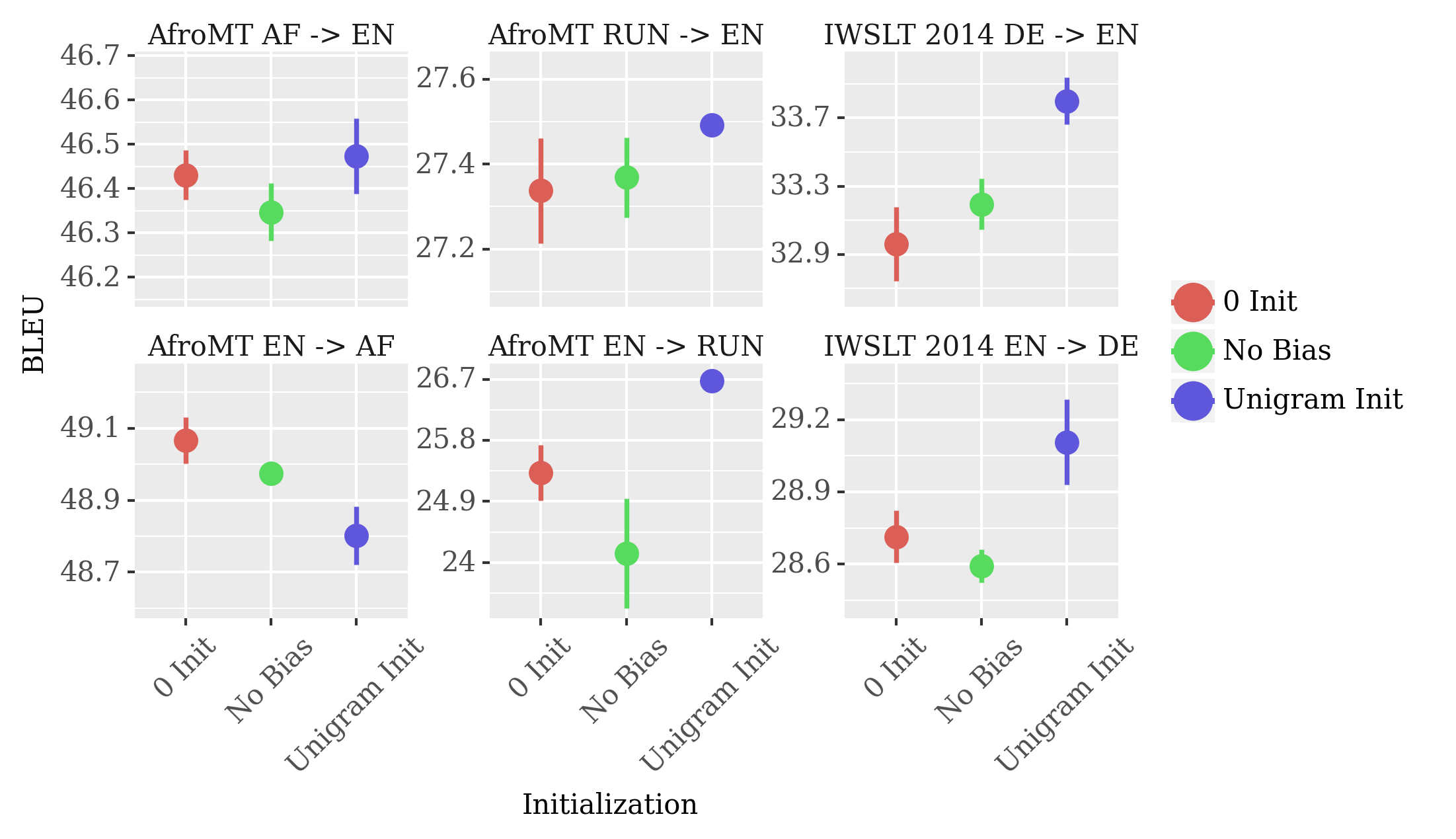}
    \caption{Mean test \bleu for models with different bias initialisations. Bars indicate standard error. Panes are adjusted so that the $y$-axis spans at least $0.5$ \bleu.}
    \label{fig:bleu}
\end{figure}
\section{Experiments}

We explore the effects of the unigram bias initialisation strategy on neural machine translation systems in comparison to a standard initialisation technique: initialising the bias to all $0$s (denoted as \zerovec) or omitting the bias term entirely.\looseness=-1 %
\subsection{Setup} 
We perform experiments with several language pairs:  WMT'14 German-to-English  \citep[De$\rightarrow$En;][]{wmt14}, IWSLT'14 German-to-English \cite[De$\leftrightarrow$En;][]{IWSLTbib}, and Afrikaans/Rundi-to-English in the AfroMT dataset \citep[Run$\leftrightarrow$En and Af$\leftrightarrow$En;][]{reid21afromt}.  
These corpora span several language families and different sizes to demonstrate performance in higher, medium and lower resource domains ($\sim\!\! 4.5$M, $\sim 750$K and $\sim 150$K sentence pairs, respectively). 
All models use the standard Transformer encoder--decoder architecture \cite{vaswani2017attention}, with 6 layers in both. The IWSLT and AfroMT Run$\leftrightarrow$En models have 4 attention heads per layer (adjusted for the smaller size of these datasets) while all other models have 8 attention heads. Dropout is set to $0.1$; the feedforward hidden dimension is set to 512 for the WMT model and 256 for all other models.  Parameter estimation is performed using stochastic gradient-descent techniques, with the standard maximum likelihood objective and label smoothing \cite{label_smoothing} with hyperparameter $\alpha=0.1$.  We use the Adam optimizer \cite{adam} with $(\beta_1, \beta_2) = (0.9, 0.997)$. Early stopping was performed during training, i.e., model parameters were taken from the checkpoint with the best validation set \bleu \cite{papineni-etal-2002-bleu}.\looseness=-1

We preprocess the data using subword tokenisation with the SentencePiece  library \cite{kudo-richardson-2018-sentencepiece}\footnote{For WMT and IWSLT, we train joint SentencePiece models with vocabulary sizes 32000 and 20480, respectively. For AfroMT, we use the SentencePiece model provided with the dataset: \url{https://github.com/machelreid/afromt}.}\tiago{SentencePiece is just a lib and not a tokenizer, right? Do we not know which was used?}
For initialisation, unigram frequencies are computed on respective training sets after tokenisation is performed. We do not hold bias term parameters fixed during 
training, although we found that 
they do not change perceptibly from their values 
at initialisation, even for the \zerovec-initialised model (\cref{app:bias_change}). The projection matrix $\bW$ in the final linear layer is initialised element-wise using $\mathcal{N}(0,\sfrac{1}{\sqrt{d}})$, where $d$ is the embedding hidden dimension;\clara{@wojciech, this is correct. The only minor difference is that instead of using normal, we have truncated normal with -2,2 as the range.}%
the matrix is then scaled such that the matrix $\ell_2$  norm is approximately equal in magnitude to the bias $\ell_2$ norm. 
 Decoding is done with length-normalised beam search with a beam size of 5, which was similarly chosen based on validation \bleu scores.  All \bleu and chrF \cite{popovic-2015-chrf} scores are computed using the sacre\bleu library \cite{post-2018-call}.

\subsection{Results}
We present main results here, and defer additional experimental results that exhibit similar trends (e.g., using chrF as the evaluation metric, or training on WMT) to \cref{app:extras}. We also explore several extensions that build on the unigram initialisation, albeit with mixed results; again, see \cref{app:extras}.\looseness=-1

\newcommand{\paragraphvspace}{-1mm}
\vspace{\paragraphvspace}
\paragraph{Performance.}
\cref{fig:bleu} presents mean test \bleu scores with standard error estimates from 5 different random seeds per dataset--intitialisation strategy combination. On 5 out of the 6 datasets, the unigram bias initialisation technique leads to comparable or better test set performance in comparison to standard bias term initialisation techniques.\looseness=-1 

\begin{figure}
    \centering
    \includegraphics[width=\columnwidth]{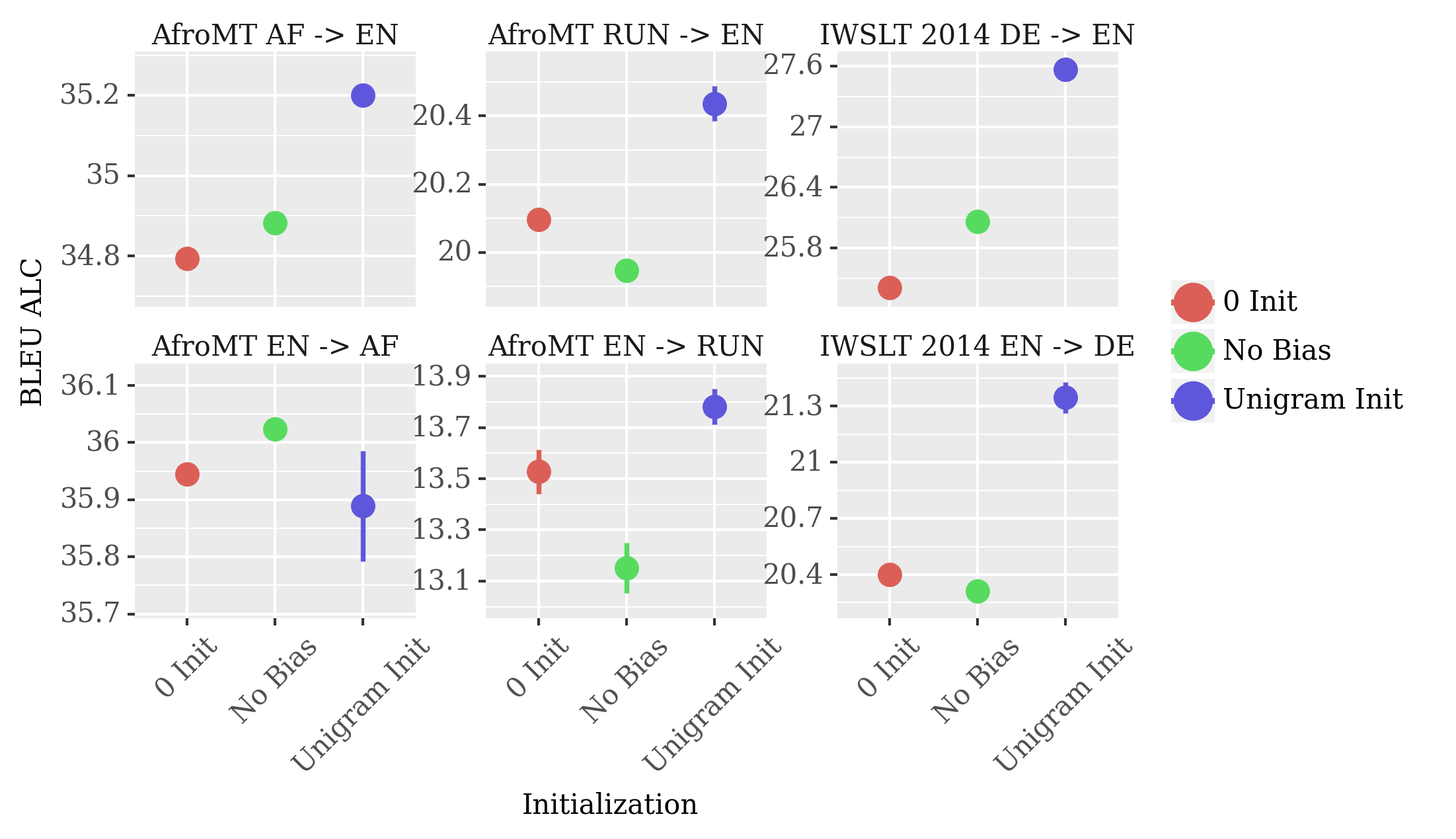}
    \caption{ALC of validation \bleu for the first $20k$ training updates for each initialisation strategy. Models initialised with the unigram distribution achieve higher \bleu earlier on in training, even when those models do not ultimately perform best (see \cref{fig:bleu}). }
    \label{fig:auc}
\end{figure}

\vspace{\paragraphvspace}
\paragraph{Efficiency.}
In order to quantify training efficiency, we estimate\footnote{Explicitly, we use the composite trapezoidal rule. } the area under the validation \bleu learning curve (ALC) \cite{pmlr-v16-guyon11a,pmlr-v123-liu20a} for the first $20$k training updates;\footnote{Models converged after $\sim 50$k updates. We look only at the first $20$k updates to focus on early training behaviours. Full training behaviours were similar (see \cref{app:training_trends}).\looseness=-1} %
for the sake of interpretability, scores are renormalised by the interval span.
From \cref{fig:auc}, we see that, on 5 out of the 6 datasets, higher \bleu is achieved earlier on in training.  Hence, the unigram bias initialisation approach appears to reach better performance in fewer iterations than standard initialisation approaches, which would be beneficial in cases where training efficiency considerations are paramount (e.g., in low-resource languages or in compute-limited settings).\looseness=-1

\begin{figure}[t]
\centering
\begin{subfigure}{0.9\linewidth}
   \includegraphics[width=1\linewidth]{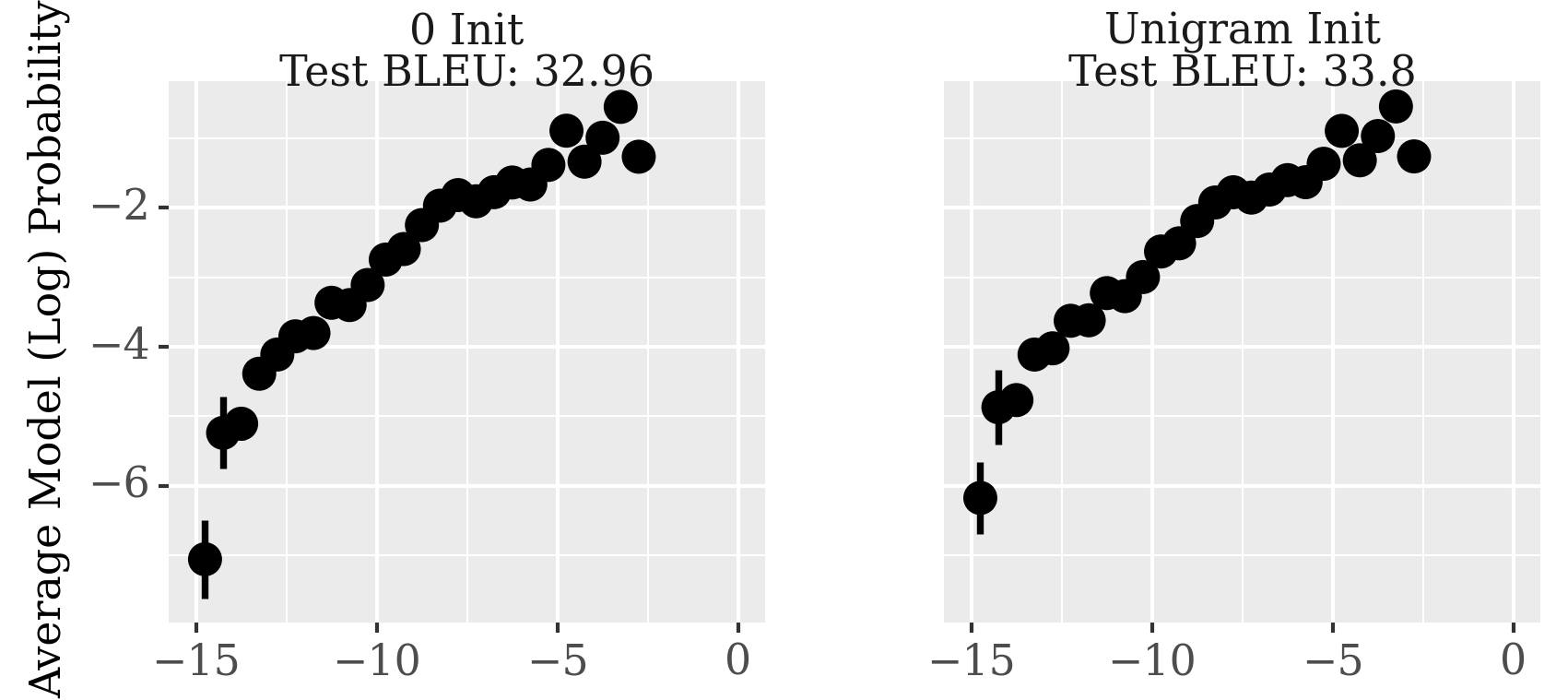}
   \caption{Full model parameters used.}\label{fig:log-probs_a}
\end{subfigure}
\vspace{-10pt}
\begin{subfigure}{0.9\linewidth}
   \includegraphics[width=1\linewidth]{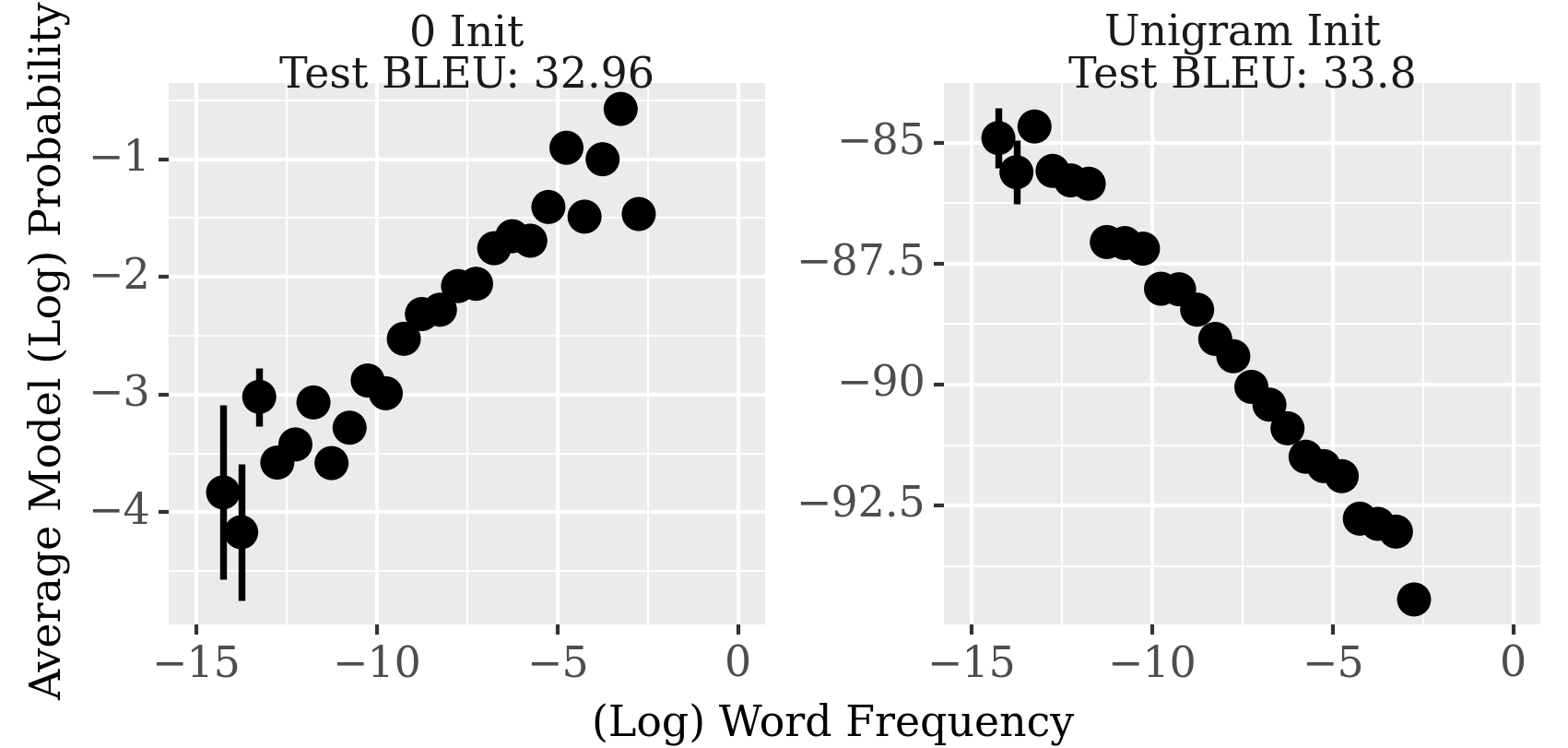}
   \caption{Bias term removed.}\label{fig:log-probs_b}
\end{subfigure}
\caption{Average log-probability assigned by the model to a ground-truth token in the test set vs. the log-unigram probability of that token in the training set (IWSLT'14 De$\rightarrow$En). Probabilities in upper row are computed using all model parameters; in the lower row, the bias term is dropped from the final linear projection.}
\label{fig:log-probs}
\end{figure}

\vspace{\paragraphvspace}
\paragraph{Analysis.}
The aim of this analysis is to investigate whether -- and to what extent -- the final-layer bias unigram initialisation leaves the contextual part of the network, $\pW(\cdot \!\mid\! \yy_{<t})$, to better capture \emph{non-frequency} effects. To this end, we examine model-assigned log-probability as a function of token frequency. 
In \cref{fig:log-probs}, we plot a token's unigram log-frequency against the average log-probability assigned to it (when it is the ground-truth token) by a model initialised with (left) a bias term of \zerovec and (right) a log-unigram bias term, binning them in equal-length intervals and averaging them for clarity. 
In \cref{fig:log-probs_a}, the full model parameters are used. In \cref{fig:log-probs_b}, the bias terms are not added in the linear projection, i.e., only the contextual part of \cref{eq:breakdown}, $\pW(\cdot \!\mid\! \yy_{<t})$, is computed. 
 
 The upward trend in average model-assigned log-probability in \cref{fig:log-probs_a} suggests that, in general, models are better (or at least more confident) when predicting more frequent tokens. This trend holds when the bias term is omitted from the final linear computation of the \zerovec-initialised model. Interestingly though, when the bias term is omitted from the unigram-initialised model, the trend appears to reverse. This change suggests that for unigram-initialised models, frequency may instead be encoded in the bias term, 
 providing evidence that for these models, $\pW(\cdot \!\mid\! \yy_{<t})$ may indeed specialise in non-frequency aspects of language.\looseness=-1 

\section{Discussion}

NLP models have been observed to overfit to surface cues in their training data, impeding their ability to generalise at inference time \cite{warstadt-etal-2020-learning,wei-etal-2021-frequency}. Thus, one could argue that learning or encoding the superficial statistical tendencies of language is not necessarily a good thing. Yet, empirical results suggest that it may in fact be an important part of model learning dynamics (see \cref{app:unigram_regularizer}, for example). Indeed, \citet{takahashi_evaluating} find evidence that more powerful language models have a natural bias for learning them. Here we ask if -- when initialising model parameters -- we can \emph{explicitly} endow our models with prior knowledge about one such statistical tendency: frequency.

While the result that this initialisation strategy improves training efficiency is perhaps not surprising, the relatively consistent improvement in overall performance \emph{is}. 
We offer two possible explanations for this improvement. The first is that this initialisation beneficially alters model learning dynamics at the beginning of training, especially as early learning dynamics can have an outsized impact on final model performance \cite{achille2018critical}. A second possible explanation is that it disentangles frequency in the modelling of contextual probabilities. %
If $p_b$ (\cref{eq:breakdown}) explicitly models the unigram distribution, then our model does not need to capture this component of the conditional distribution in its other parameters, which frees up model capacity to focus on more complex phenomena within natural language. Its success thus motivates exploring the use of higher-order statistical models,  such as a bigram or trigram model, in an attempt to further disentangle  surfaces statistics from more nuanced components of natural language in a modular fashion.\looseness=-1

\section{Conclusion and Future Work}
In this work, we revisit a simple initialisation technique in the context of modern neural language generation models: setting the bias term in the final linear projection layer to the log-unigram distribution of (sub)words within the training corpus. This strategy leads to more efficient training; perhaps more surprisingly, it also leads to better overall performance in our machine translation experiments. We offer analysis and discussion as to the cause of these trends.
An interesting direction for future work could be determining the effects that this initialisation procedure has on various model properties,  e.g., its embedding space, and its benefits specifically in low-resource settings. Furthermore, extensions of this work could explore potential uses of this strategy in the mitigation of problems with lexically infrequent words, e.g., by analysing via the decomposition in \cref{eq:breakdown} whether a model's probability estimate for a word is being driven by frequency or contextual components. Finally, this technique is not limited to models of distributions over strings; it is in fact applicable to \emph{any} neural classification setting, the exploration of which is left to future work.%
\section{Acknowledgements}
We would like to thank the members of the DeepMind Language Team for insightful discussions during the course of this work and specifically, Chris Dyer and Kris Cao for helpful  feedback on the initial version of this paper and John Hale for pointers to references on human language acquisition. We would also like to thank Clément Guerner for detailed feedback on clarity and presentation. 

\section{Limitations}
Perhaps the main limitation of this work is that we only explore the approach within the context of machine translation benchmarks, although we conduct extensive experiments within this task that cover different training data scales and diverse pairs of languages, including low-resource ones. Nevertheless, we remark that the proposed approach is entirely general-purpose, and can be applied to \emph{any} other language generation or even any neural classification tasks. We leave it to future work to investigate whether the same gains would apply in those settings.  
Furthermore, we have not yet explored how this technique would interact with other modelling choices, such as different optimizers, training objectives, or subword tokenisation algorithms. Lastly, our unigram initialisation of the bias term is currently done at the level of subword units, which do not always correspond to lexically or morphologically meaningful linguistic units. We leave the extension of this approach to more meaningful linguistic units, such as words or morphemes, to future work.%

\section{Ethical Considerations}
We foresee no ethical issues that could arise from the findings presented in this work.

\bibliography{anthology,custom}
\bibliographystyle{acl_natbib}

\clearpage
\newpage
\appendix

\section{Additional Experiments}\label{app:extras}
\subsection{Additional Training Trends}\label{app:training_trends}
\begin{figure}[h]
    \centering
    \includegraphics[width=\columnwidth]{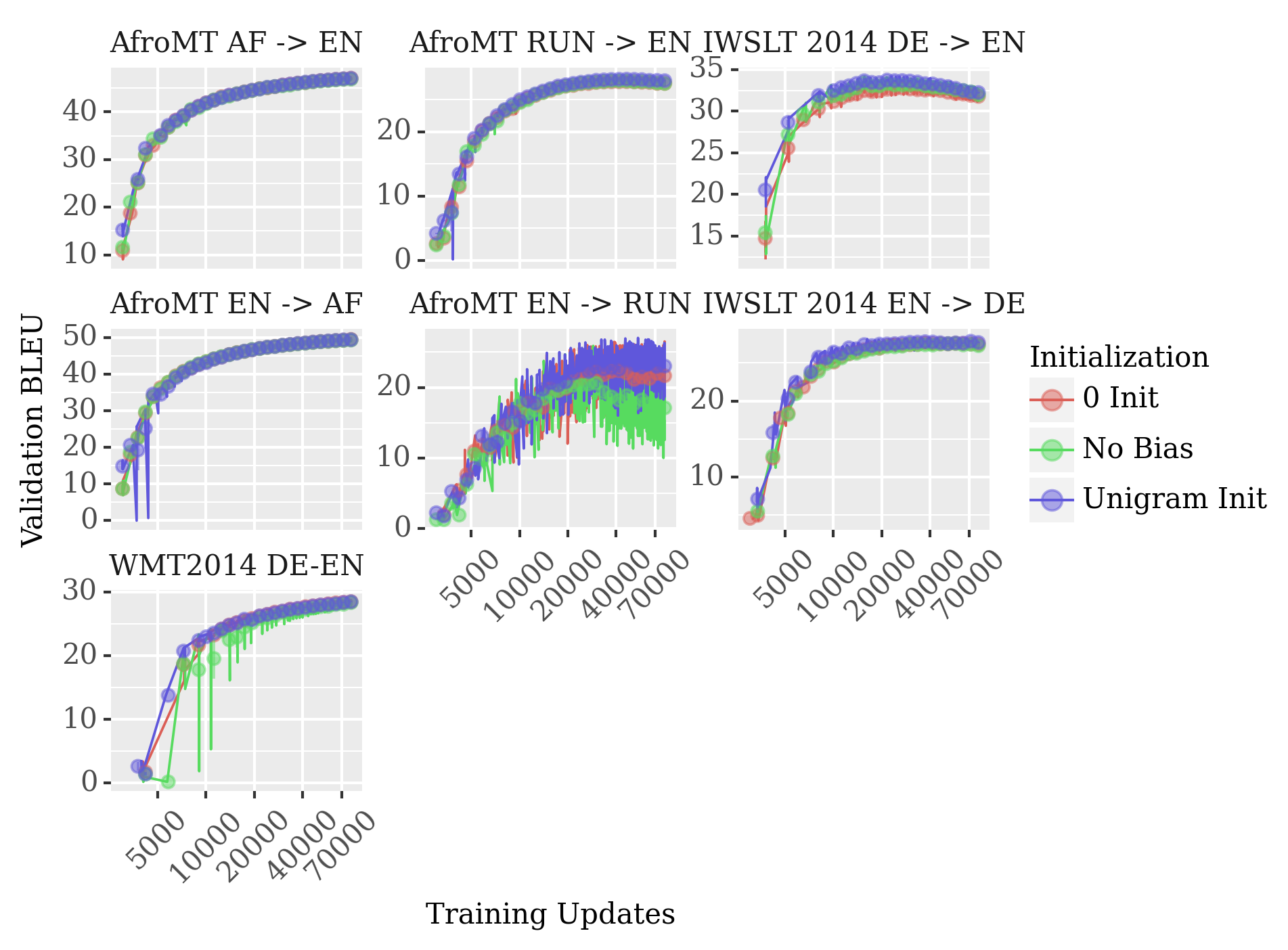}
    \caption{Validation BLEU over the course of training. $y$-axis is on log-scale.}
    \label{fig:bleu_over_training}
\end{figure}
\begin{figure}[h]
    \centering
    \includegraphics[width=\columnwidth]{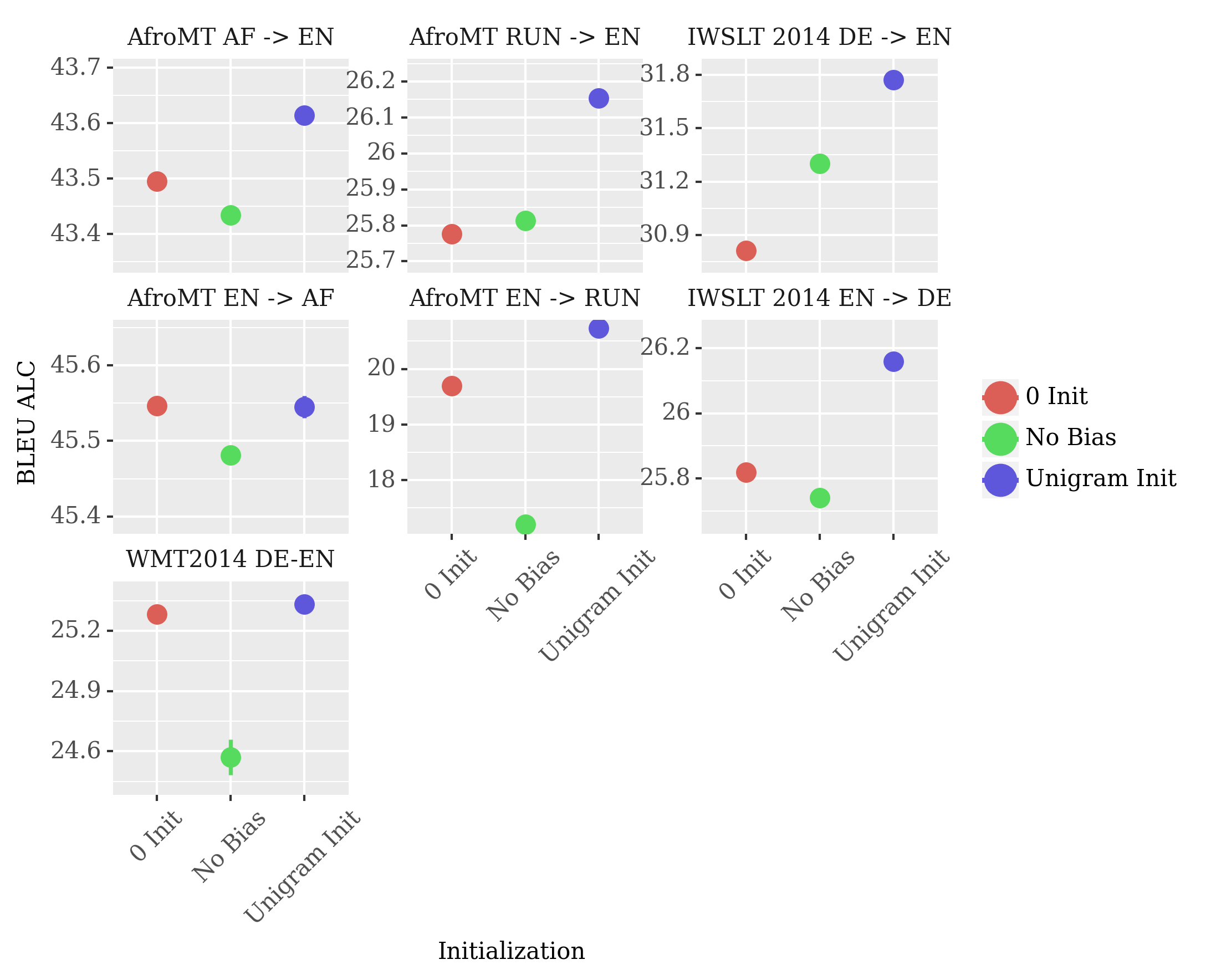}
    \caption{ALC scores for models over the full course of training. All other experimental details are the same as in \cref{fig:auc}.}
    \label{fig:full_auc}
\end{figure}

\subsection{WMT Experiments}
\begin{figure}[h]
    \centering
    \includegraphics[width=\columnwidth]{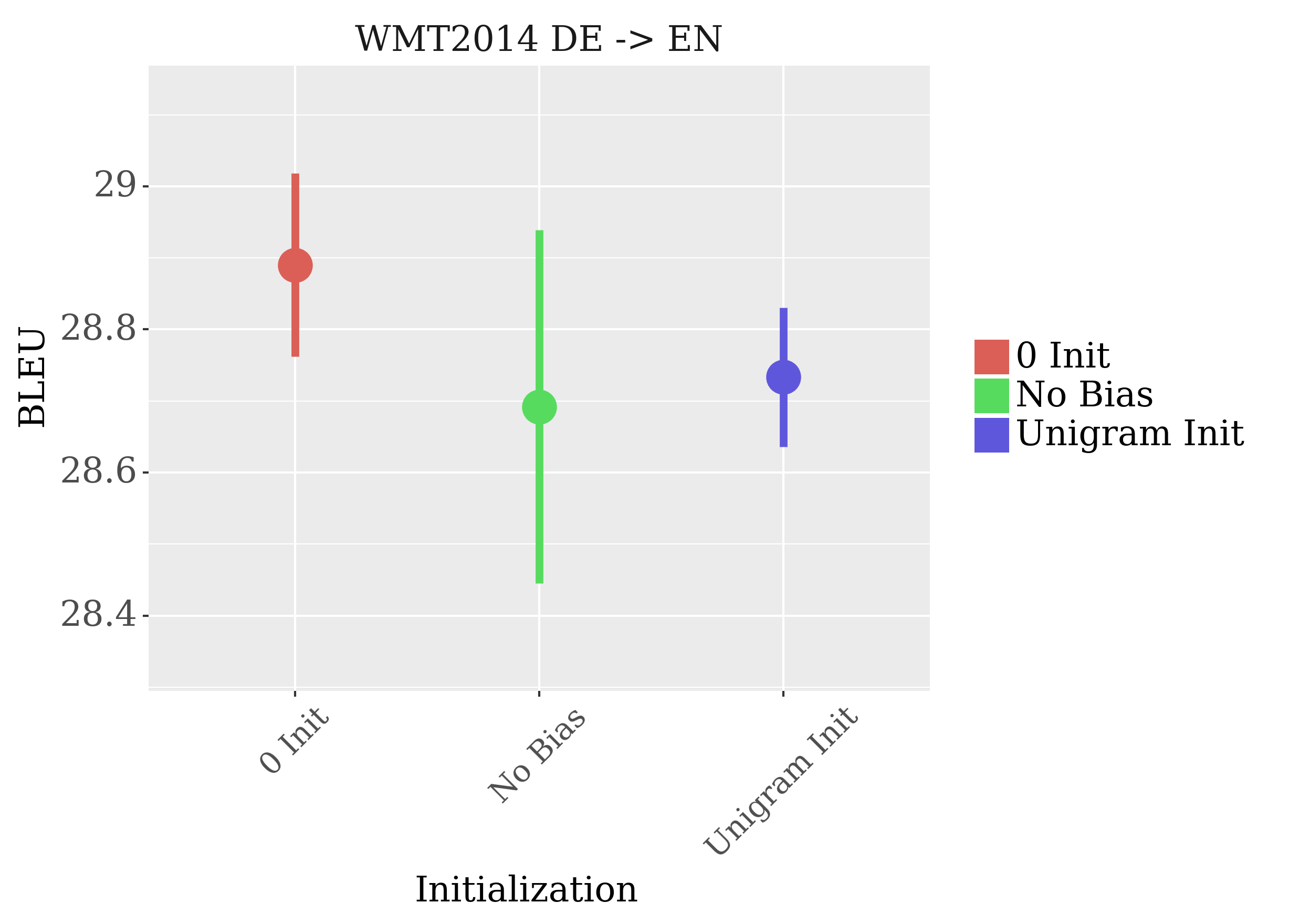}
    \caption{\bleu scores on WMT'14 De$\rightarrow$En. Setup is the same as in \cref{fig:bleu}.}
    \label{fig:wmt_bleu}
\end{figure}

\begin{figure}[h]
    \centering
    \includegraphics[width=\columnwidth]{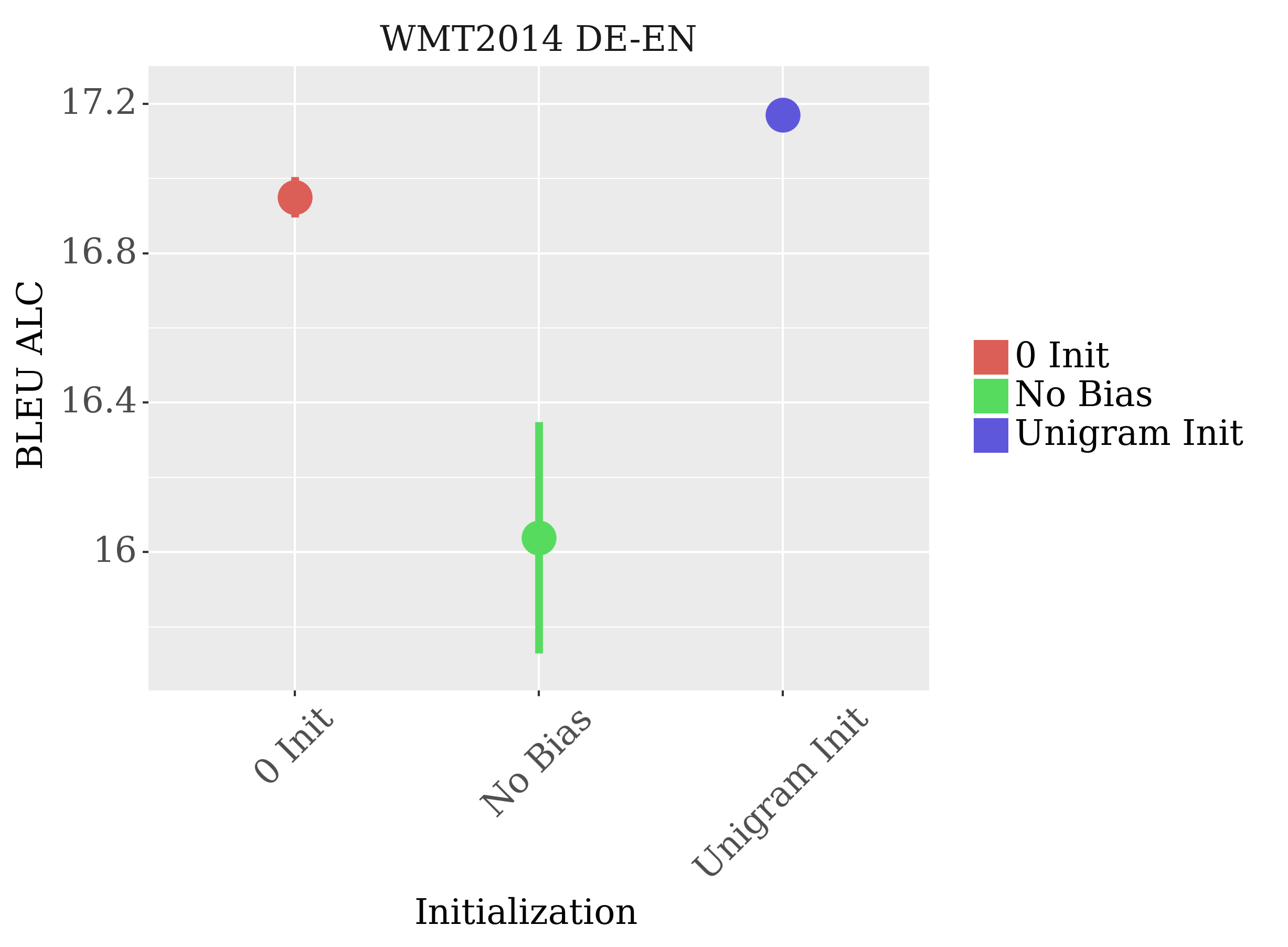}
    \caption{ALC scores on WMT'14 De$\rightarrow$En. Setup is the same as in \cref{fig:auc}.}
    \label{fig:wmt_auc}
\end{figure}

\subsection{chrF Scores}

\begin{figure}[H]
    \centering
    \includegraphics[width=\columnwidth]{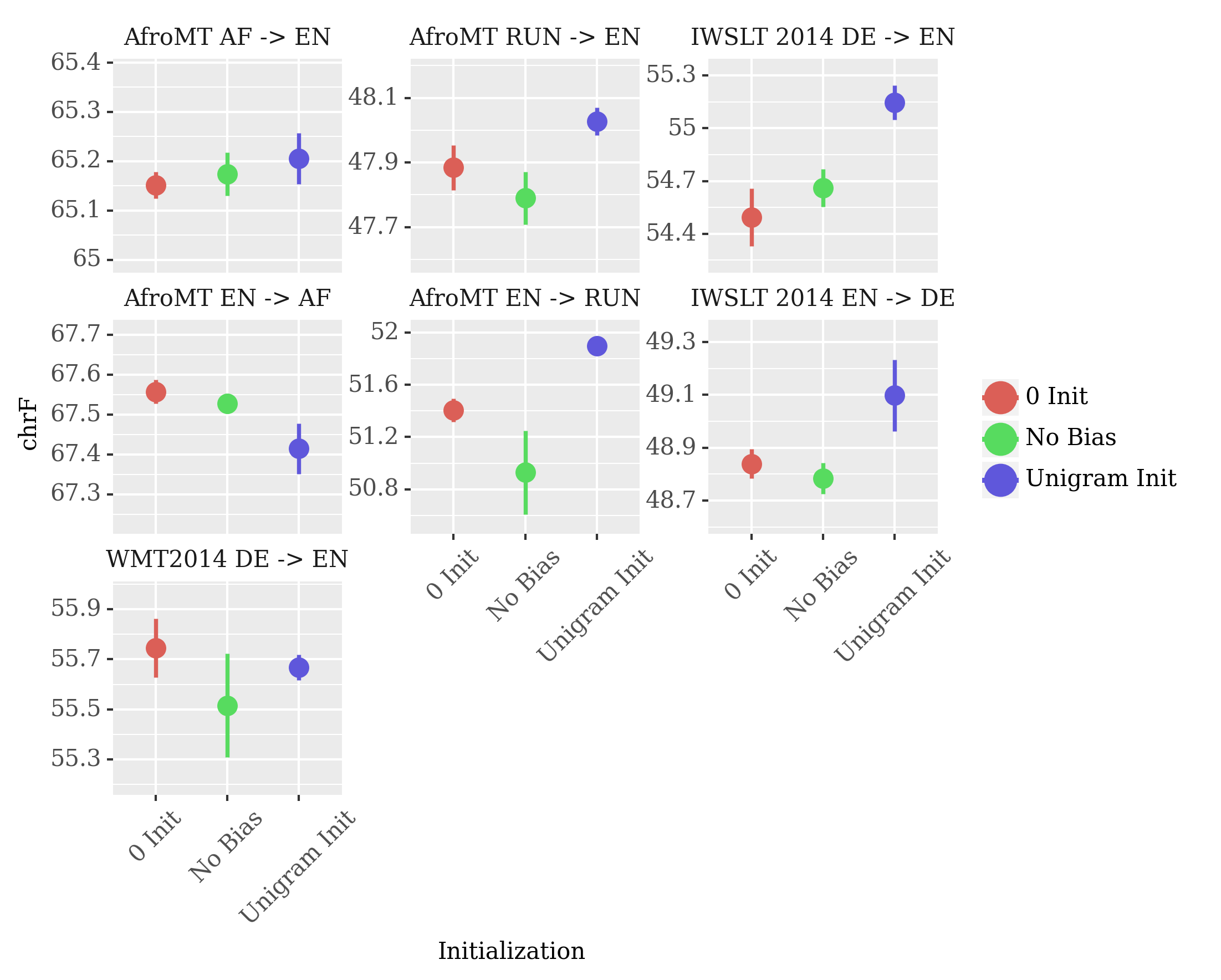}
    \caption{chrF scores on test set for models in \cref{fig:bleu}. We observe the same trends.}
    \label{fig:chrf}
\end{figure}
\begin{figure}[H]
    \centering
    \includegraphics[width=\columnwidth]{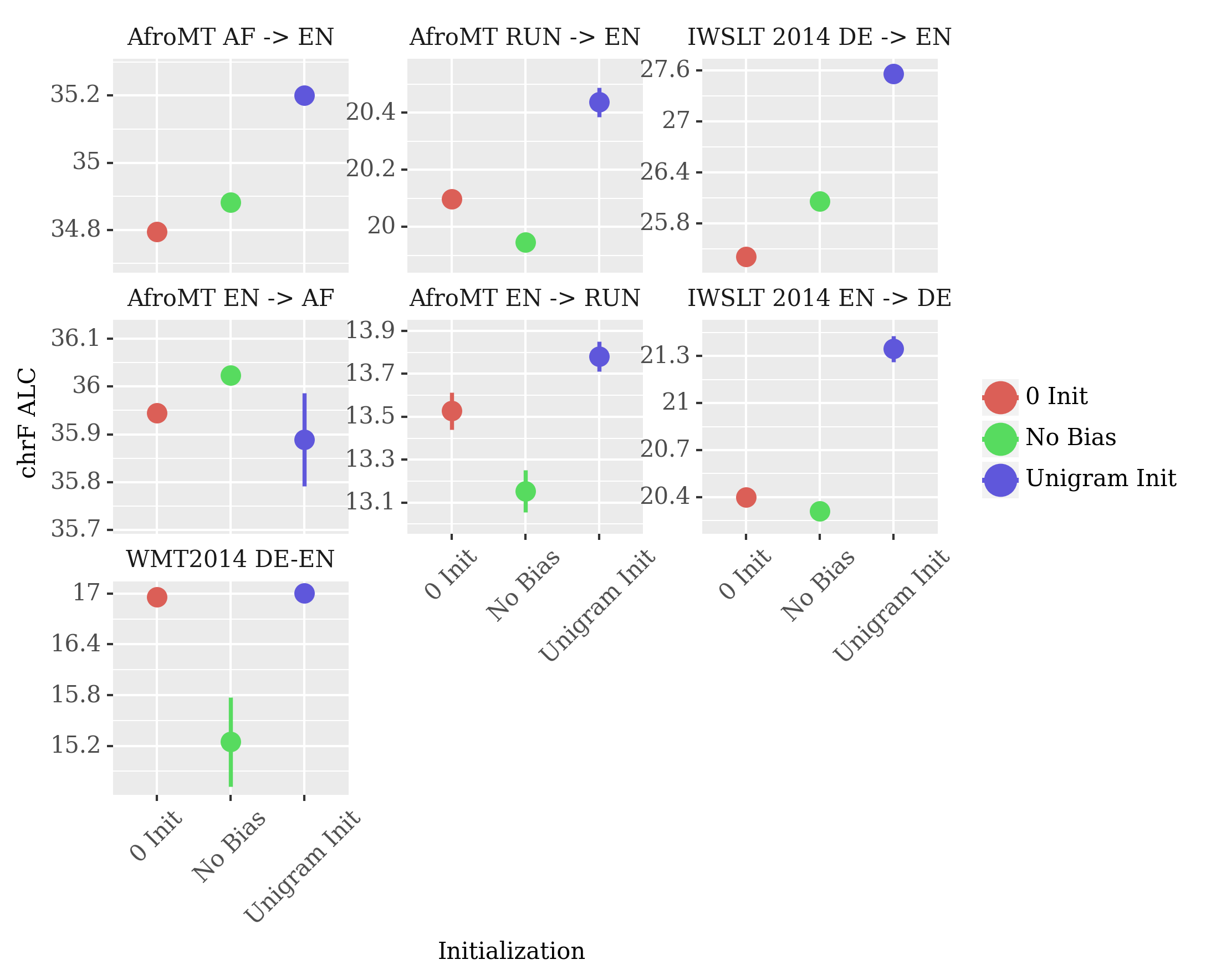}
    \caption{ALC scores using chrF as metric; again only the first $20$k steps are considered.}
    \label{fig:chrf_auc}
\end{figure}
\subsection{Regularising Away from the Unigram Distribution}\label{app:unigram_regularizer}
\begin{figure}[H]
    \centering
    \includegraphics[width=\columnwidth]{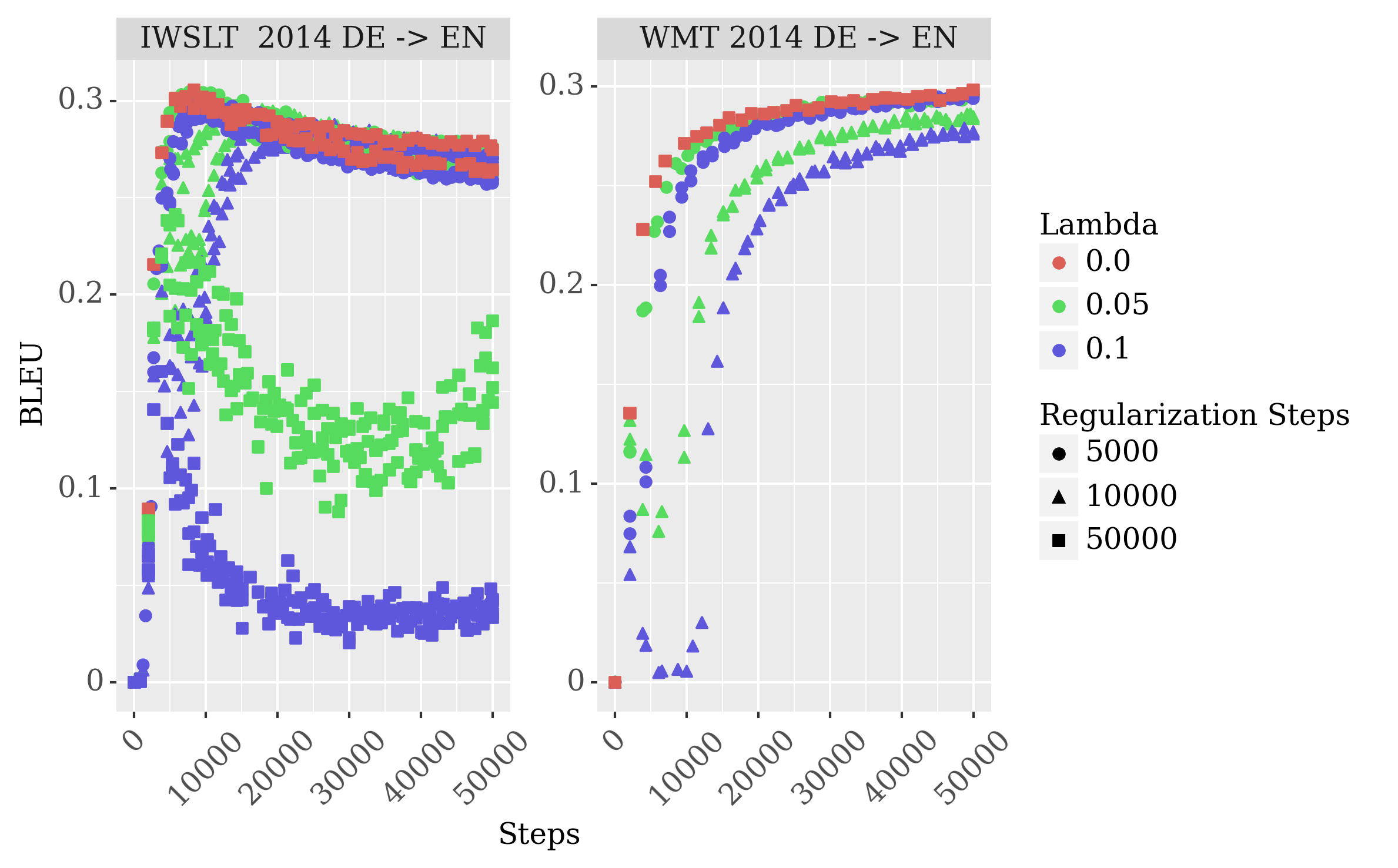}
    \caption{\bleu scores (in decimal format)  when regularising away from the unigram distribution. ``Regularisation steps'' indicates the point at which we change back to the standard objective.}
    \label{fig:reg}
\end{figure}

Prior work has suggested that models' learning of surface statistics, such as the unigram distribution, may harm their generalisation abilities \cite{warstadt-etal-2020-learning,wei-etal-2021-frequency}. Under this premise, it seems feasible that the learning trends observed in \cref{fig:unigram_div} could have downstream negative side-effects, e.g., the inappropriate preference for higher frequency words observed in \cite{wei-etal-2021-frequency}. Given the importance of early stage training dynamics \cite{achille2018critical}, it may even be the root cause of such behaviour.  In the effort to test this hypothesis, we try to regularise a model's output \emph{away} from the unigram distribution in early stages of training. Specifically, we instead minimise the objective $\mathrm{KL}(p \mid\mid  \model) - \lambda\,\mathrm{KL}(\omega(p) \mid\mid  \model)$ for empirical distribution $p$ and the unigram distribution of this empirical distribution $\omega(p)$. $\lambda$ is a hyperparameter. We use this objective for the initial steps of training, then switching back to the standard objective $\mathrm{KL}(p \mid\mid  \model)$. In \cref{fig:reg}, we observe that this form of regularisation leads to worse (or equivalently performing) models by the time of convergence. Results were similar when evaluated on out-of-distribution data.
\subsection{Out-of-Domain Performance}
\begin{figure}[H]
    \centering
    \includegraphics[width=\columnwidth]{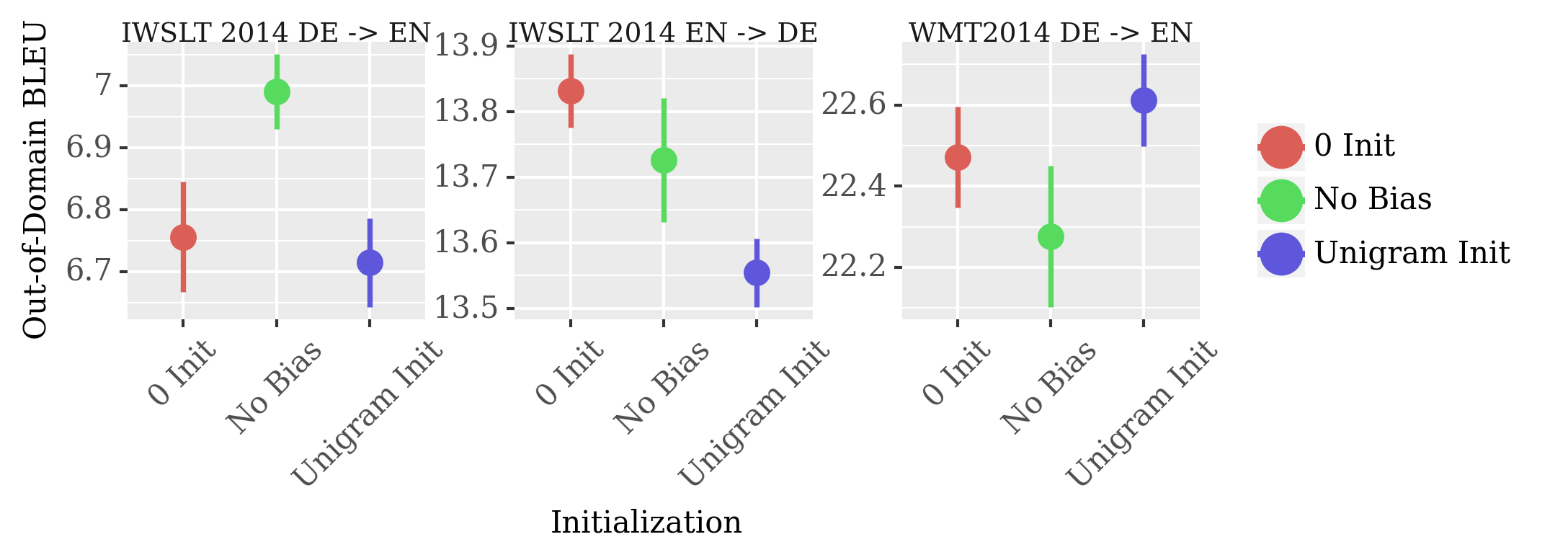}
    \caption{\bleu on out-of-domain test sets (WMT14 for IWSLT and vice-versa) by bias initialisation strategies. Bars indicate standard error. }
    \label{fig:ood_bleu}
\end{figure}
\subsection{Change in Bias Term over Training}\label{app:bias_change} 
\begin{figure}[H]
    \centering
    \includegraphics[width=\columnwidth]{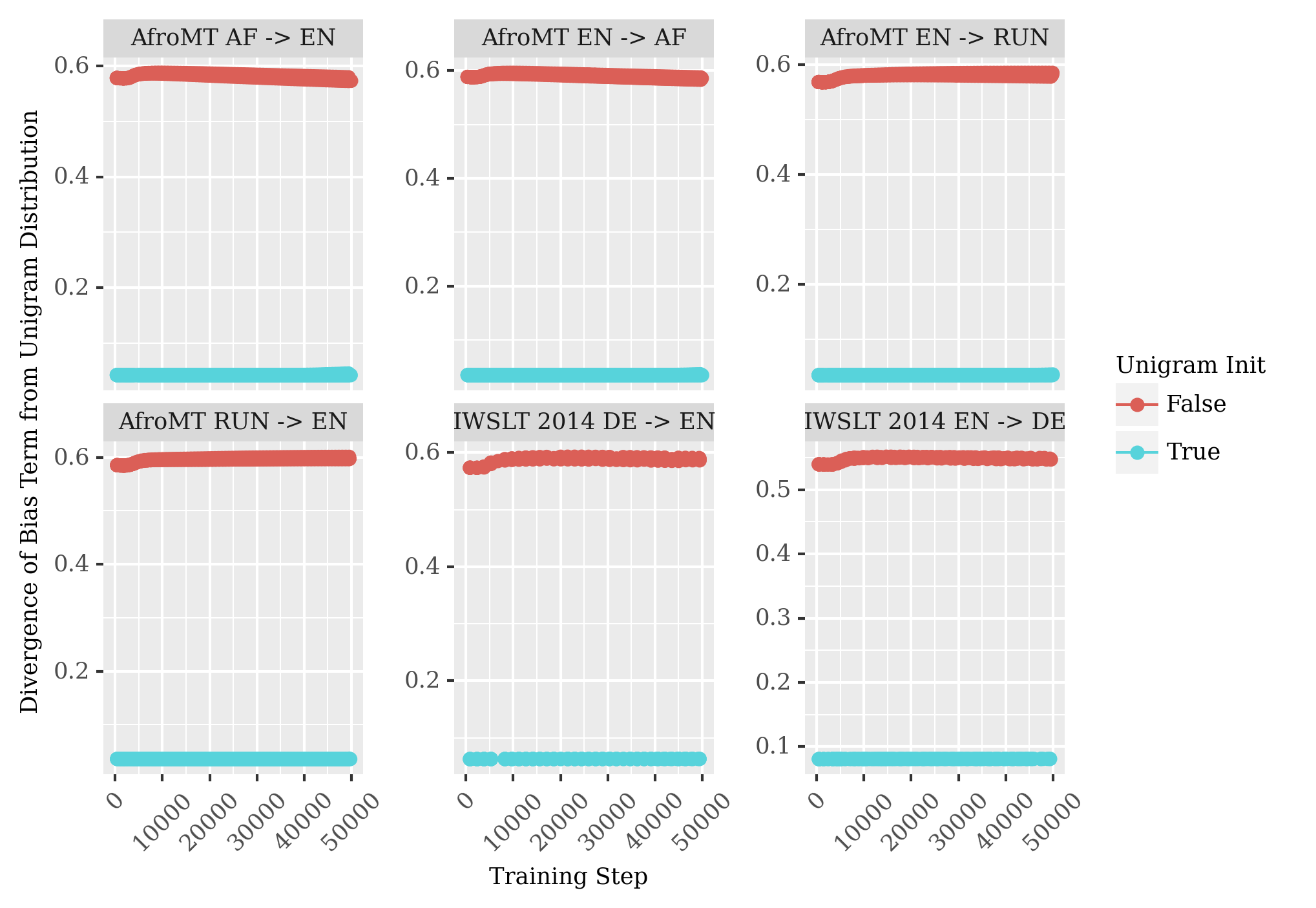}
    \caption{Divergence of the bias term (after $\mathrm{softmax}$ is performed to map bias onto probability simplex) from the unigram distribution of the respective training set for models trained on different data sets. Same models as used in \cref{fig:auc}. }
    \label{fig:unigram_bias_div}
\end{figure}

\begin{figure}[H]
    \centering
    \includegraphics[width=\columnwidth]{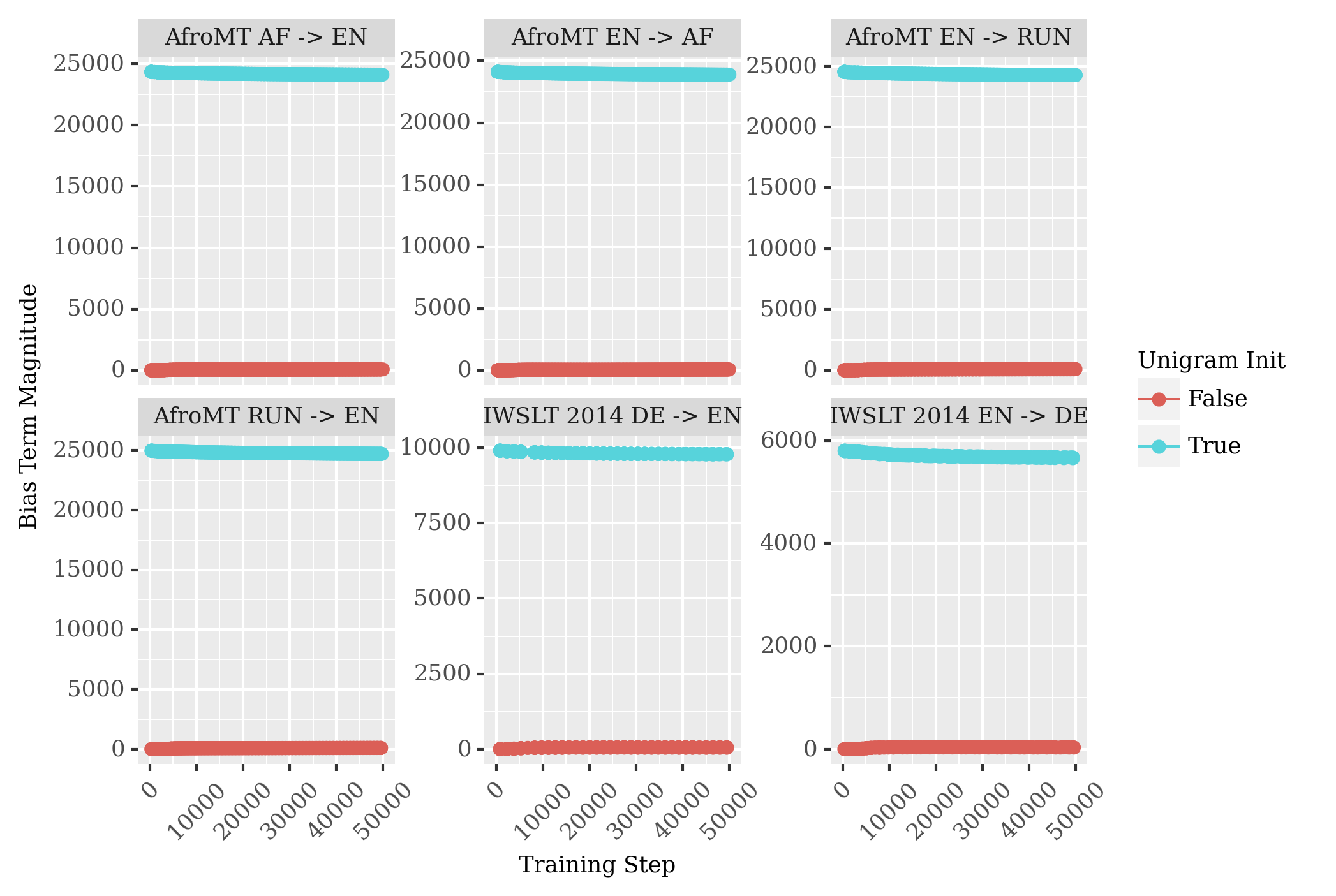}
    \caption{Magnitude of the bias term in models trained on different data sets. Same models as used in \cref{fig:auc}. Along with \cref{fig:unigram_bias_div}, these results suggest that the bias term does not change much from its value at parameter initialisation. }
    \label{fig:unigram_bias_mag}
\end{figure}

In \cref{fig:unigram_bias_div,fig:unigram_bias_mag}, we see the divergence of the bias term from the unigram distribution and the magnitude of the bias term, respectively. Interestingly, we see that neither value changes perceptibly from the time of initialisation onward, suggesting the bias term itself does not change much from its initialised value. This trend is consistent across seeds and datasets.

\subsection{Initialisation with Bias Term from Large-Scale Dataset}
We additionally explore the effects of initialising the bias term with the log-unigram distribution, as estimated from a larger dataset in a more general purpose domain. We hypothesise that this strategy could be useful in low resource settings.  We find that this indeed improves the generalisation performance of a model trained on IWSLT when evaluated on an OOD dataset (see \cref{fig:c4_init_ood}).
\begin{figure}[H]
    \centering
    \includegraphics[width=\columnwidth]{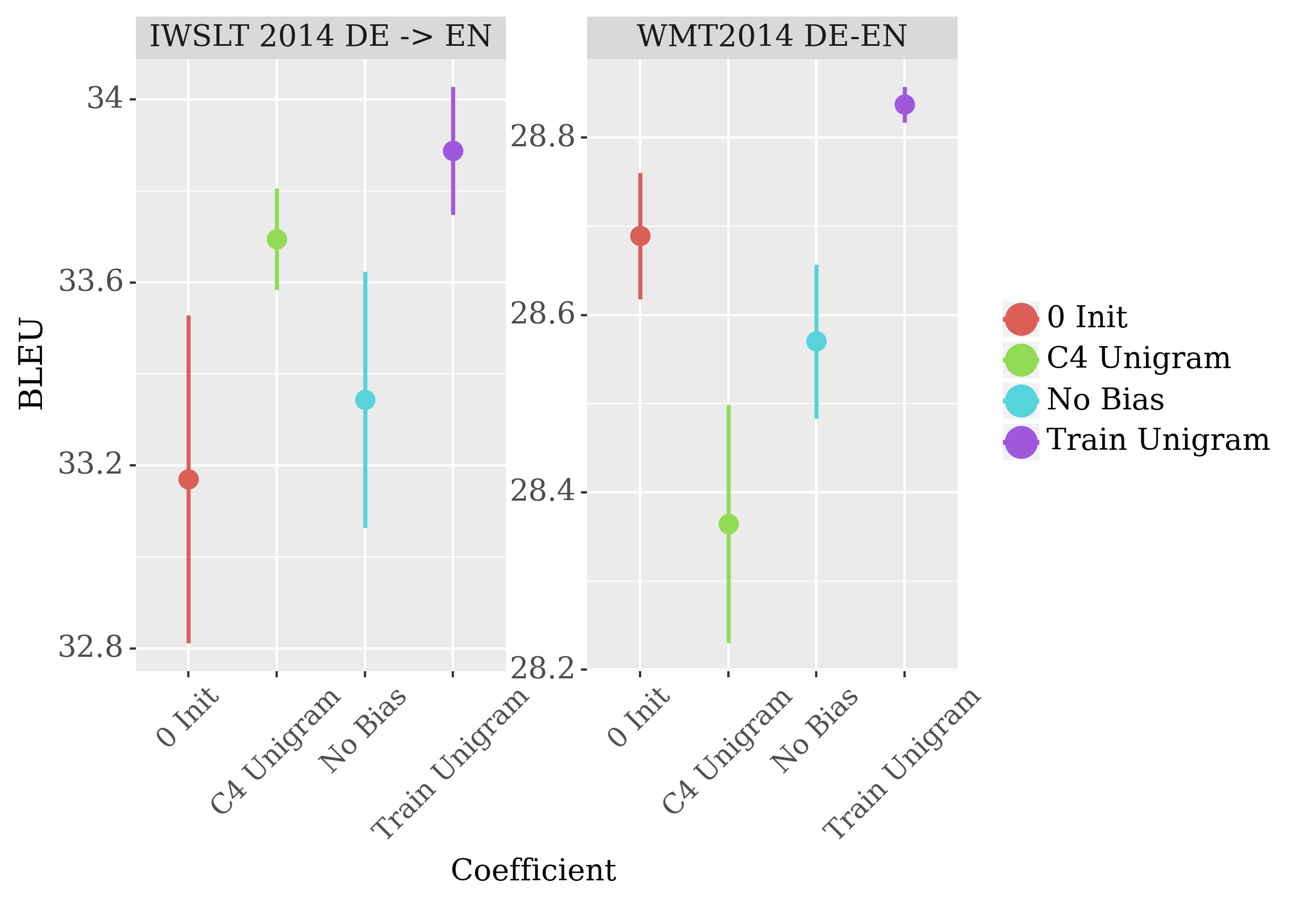}
    \caption{Results of best model checkpoint on validation set of respective datasets. Models are initialised with a \zerovec bias term, no bias term, or the log-unigram distribution of either the C4 English dataset \cite{2019t5} or the respective training dataset.}
    \label{fig:c4_init}
\end{figure}
\begin{figure}[H]
    \centering
    \includegraphics[width=\columnwidth]{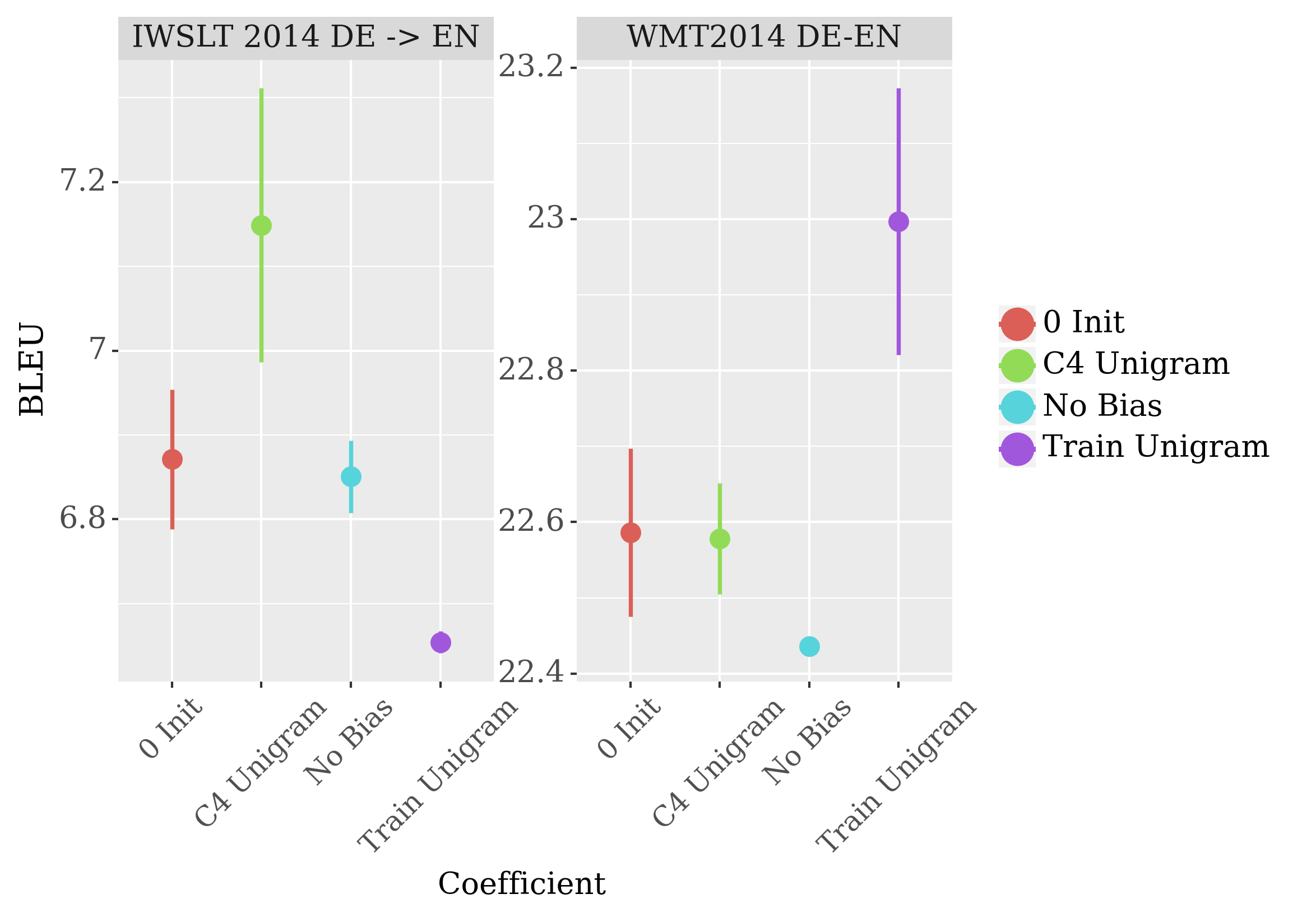}
    \caption{Results of best model checkpoint trained on IWSLT and WMT (as specified in facet label) and evaluated on (validation set of) out-of-domain datasets (WMT for the IWSLT model and vice-versa). Models are same as those in \cref{fig:c4_init}.}
    \label{fig:c4_init_ood}
\end{figure}

\end{document}